\documentclass[journal]{IEEEtran}
\usepackage[usenames,dvipsnames,table]{xcolor}

\usepackage{amsmath}
\usepackage{amssymb}
\usepackage{booktabs}
\usepackage{multirow}
\usepackage[ruled,lined,linesnumbered]{algorithm2e}
\usepackage{balance}
\usepackage{amsmath,graphicx}
\usepackage{subfigure}
\usepackage{tabularx}
\usepackage{rotating}
\usepackage[lofdepth,lotdepth]{subfig}
\usepackage{pifont} 

\graphicspath{{Figures/}}
\ifCLASSINFOpdf
\else

\fi

\begin{document}
\title{Human Activity Recognition Based on Wearable Sensor Data: A Standardization of the State-of-the-Art}

\author{Artur Jordao,
        Antonio Carlos Nazare,
        Jessica Sena,
       William Robson Schwartz
       \\
       Smart Sense Laboratory, Computer Science Department, 
       Universidade Federal de Minas Gerais, Brazil
       \\
       Email: \{arturjordao, antonio.nazare, jessicasena, william\}@dcc.ufmg.br}

\maketitle

\begin{abstract}
Human activity recognition based on wearable sensor data has been an attractive research topic due to its application in areas such as healthcare and smart environments. In this context, many works have presented remarkable results using accelerometer, gyroscope and magnetometer data to represent the activities categories. However, current studies do not consider important issues that lead to skewed results, making it hard to assess the quality of sensor-based human activity recognition and preventing a direct comparison of previous works. These issues include the samples generation processes and the validation protocols used. We emphasize that in other research areas, such as image classification and object detection, these issues are already well-defined, which brings more efforts towards the application. Inspired by this, we conduct an extensive set of experiments that analyze different sample generation processes and validation protocols to indicate the vulnerable points in human activity recognition based on wearable sensor data. For this purpose, we implement and evaluate several top-performance methods, ranging from handcrafted-based approaches to convolutional neural networks. According to our study, most of the experimental evaluations that are currently employed are not adequate to perform the activity recognition in the context of wearable sensor data, in which the recognition accuracy drops considerably when compared to an appropriate evaluation approach. To the best of our knowledge, this is the first study that tackles essential issues that compromise the understanding of the performance in human activity recognition based on wearable sensor data.
\end{abstract}

\section{Introduction}\label{sec::introduction}
Due to technological advances, activity recognition based on wearable sensors has attracted a large number of studies. This task consists of assigning a category of activity to the signal provided by wearable sensors such as accelerometer, gyroscope and magnetometer. With this purpose, there are two types of approaches. One investigates how to represent the raw signal to better distinguish the activities~\cite{Khan:2010, Bayat:2014} and the other explores the classification stage, which might consider either the raw signal~\cite{Chen:2015,Ha:2015,Ha:2016,Zeng:2014}, a pre-defined representation~\cite{Kwapisz:2010, Catal:2015, Kim:2012, Kim:2013} or a combination of both~\cite{Jiang:2015, Jordao:2018}.
Even though many improvements have been achieved in both approaches~\cite{Lara:2013,Shoaib:2015,Wang:2017}, it is hard to assess the quality of  sensor-based human activity recognition since previous studies have not contemplated the following issues.

\begin{enumerate}
\item Since many works propose their own datasets to conduct the evaluation, it is unclear which datasets are more adequate. In addition, these datasets are not always publicly available, preventing the reproducibility of the work.
\item The metrics and the validation protocol employed to assess the activity recognition quality vary from paper to paper (e.g., the {third and fourth columns} in Table~\ref{tab::protocols}).
%
%
\item The process to generate the data samples, before performing the evaluation, presents a wide variability (e.g., the second column in Table~\ref{tab::protocols}).
\end{enumerate}
While the first two issues play an important role, the last is a critical point, influencing the final performance of the activity recognition. It is important to emphasize that in other research areas, such as object detection~\cite{VOC:2007}, image classification~\cite{ImageNet:2015} and face verification~\cite{LFW:2007, YTF:2011}, these issues are handled, leading to a standardized evaluation, which attracts more research efforts towards the application. 
As a consequence of these issues, currently it is not possible to know the {state-of-the-art} methods in human activity recognition based on wearable sensor data. Additionally, since the performance of the methods can be skewed by the conducted evaluation, it is difficult to identify promising research directions.

The aforementioned discussion motivated our study, in which we implement several existing works and evaluate them on a large set of publicity available datasets. To provide a more robust evaluation and show statistical equivalence between the methods, we also perform a statistical validation~\cite{Raj:1990}.

The development of this work presents the following contributions: 
1) implementation and evaluation of several top-performance methods to human activity recognition based on wearable sensors, ranging from handcrafted-based to convolutional neural networks approaches;
2) demonstration that the process chosen to generate the data samples is a crucial point since all methods have their recognition accuracy reduced when evaluated on a unbiased sample generation process;
3) proposition of two novel data sample generation processes: \emph{Full-Non-Overlapping-Window} and \emph{Leave-One-Trial-Out}, in  which each one handles a particular drawback of the currently employed sample generation processes; and
4) standardization of  popular datasets focused on human activity recognition associated with wearable sensor data to facilitate their use and the evaluation of future works.

Since nowadays there are many works in human activity recognition based on wearable sensors, we selected the ones that provide enough information to reproducibility (i.e., the definition of the features employed and the classifier setup). In addition, regarding the datasets employed in our study, we select those used by previous works~\cite{Kim:2013, Jiang:2015, Catal:2015, Ha:2015, Ha:2016}. These datasets vary in the number of activities, sampling rate and types of sensors. This way, it is possible to examine the robustness of the methods on data with high variability.

The experimental results show that the currently used process to generate the data samples is not adequate to assess the quality of the activity recognition  since current methods allow a data sample (or part of its content) appear in both training and testing (details in Section~\ref{sec::protocol}), thereby, when appropriate data sample generation techniques are employed, the accuracy drops, on average, ten percentage points.
Therefore, the results reported by previous works can be skewed and might not reflect their real performance.

To the best of our knowledge, this is the first study that tackles essential issues that compromise the results achieved in sensor-based human recognition, such issues have been responsible for skewing some previous results reported in the literature.
%
%
We hope that this work allows a better comprehension of the challenges in human activity recognition based on wearable data and leads to further advancements of future works.
All results of this work, including the data and implementations to reproduce them, are available\footnote{http://www.sense.dcc.ufmg.br/activity-recognition-based-wearable-sensors/}.
\section{Human Activity Recognition based on Wearable Sensor Data}\label{sec::relatedwork}
This section starts by describing some surveys related to the progress in human activity recognition based on wearable sensor data. Then, it introduces details regarding the works evaluated in our study, where we discuss approaches based on handcrafted and convolutional neural network. 

\subsection{Literature Reviews} 
One of the most comprehensive studies in human activity recognition based on wearable sensors is the work of Shoaib et al.~\cite{Shoaib:2015}. Their work describes limitations and recommendations to online activity recognition using mobile phones. The term online refers to the implementation of the complete classification pipeline on the mobile phone, which consists on describing and classifying the signal. However, their work does not take into account convolutional neural network approaches, which nowadays are the most employed methods~\cite{Ha:2016, Wang:2017}.
On the other hand, Wang et al.~\cite{Wang:2017} performed an extensive study regarding these approaches in the context of wearable sensors. In their work, the authors survey a number of deep learning based methods, including recurrent neural networks (RNN) and stacked autoencoders. 

Stisen et al.~\cite{Stisen:2015} investigated the influence of heterogeneous devices on the final performance of the classifiers to to perform activity recognition. For this purpose, the authors represented the activities using handcrafted features and employed popular classifiers such as nearest-neighbor, support vector machines and random forest. Additionally, in their work, the authors noticed that severe sampling instabilities occur in the devices, which contributes to a more challenging activity recognition.
Similar to~\cite{Shoaib:2015, Stisen:2015, Wang:2017}, Mukhopadhyay~\cite{Mukhopadhyay:2015} performed a detailed investigation regarding the advances in activity recognition associated with inertial data; however, focusing on the hardware context.

Different from these works, we do not summarize or review existing methods based on their reported results. Instead, we implement and conduct an extensive set of experiments on the methods to show crucial questions that affect them.
\subsection{Methods based on Handcrafted Features}
To represent the activities, Kwapisz et al.~\cite{Kwapisz:2010} extracted handcrafted features (i.e., average and standard deviation) from the raw signal. The authors also analyzed a set of classifiers to determinate the best one able to classify the categories of activities. With this purpose, Kwapisz et al.~\cite{Kwapisz:2010} examined multilayer perceptron, decision tree (J48) and logistic regression, where the first achieved the best classification results.
Following these ideas, Catal et al.~\cite{Catal:2015} proposed to apply ensemble techniques to combine these classifiers and compose the final predictor. By performing this process on the same features, Catal et al.~\cite{Catal:2015} achieved a more accurate classification when compared with Kwapisz et al.~\cite{Kwapisz:2010}.

To increase the discriminative power among activities, Kim et al.~\cite{Kim:2012} divided an entire activity (in our context, a sample provided by temporal window process, details in Section~\ref{sec::exp_setup}) into a set of action units. Each action unit is represented by its average and correlation and is classified using bagging of decision trees. Finally, based on the proportion of each action unit, it is  possible to predict which activity these action units belong. In contrast to~\cite{Kim:2012}, Kim et al.~\cite{Kim:2013} proposed to use the boosting (compose of decision trees) with a smaller number of action units. 
%
The main difference between~\cite{Kim:2012} and~\cite{Kim:2013} is the number of action units and the classifier employed. Therefore, in this work, we only report the activity recognition accuracy of~\cite{Kim:2012}, which presented the best results in our experiments. Additionally, since the majority of the datasets do not provide enough information to build the activity units, we examine the work of~\cite{Kim:2012} in terms of the features and classifier.

It is important to emphasize that some of the aforementioned works evaluate many classifiers. This way, to reduce the number of experiments and standardize their methods, our implementation considers only the best classifier of each method, according to their original paper.
\subsection{Methods based on Convolutional Neural Networks}
Another increasing line of research in human activity recognition based on wearable sensors aims at avoiding the design of handcrafted features, operation that requires human work and expert knowledge. These works employ convolutional neural networks (ConvNet) to learn the features and the classifier simultaneously.

Focusing on convolutional neural networks, Chen and Xue~\cite{Chen:2015} employed a sophisticated ConvNet, where the input is taken from the raw signal (details in Section~\ref{sec::exp_setup}). They proposed a ConvNet architecture composed of three convolutional layers with $18, 36, 24$ filters, respectively, followed by $2\times1$ max-pooling layers, each. To extract the association between two neighboring pairs of signal axes, at the first layer, Chen and Xue applied a $12\times2$ (height $\times$ width) convolutional kernel, while in the remaining layers the authors capture only the temporal relation with kernels $12\times1$. 

Similarly to~\cite{Chen:2015}, Jiang and Yin~\cite{Jiang:2015} introduced a ConvNet of two layers with convolutional kernels of $5\times5$ followed by $4\times4$ and $2\times2$ average-pooling layers, respectively. To improve the representation of the input data before presenting them to the ConvNet, the authors perform a process, referred to as \emph{signal image}, which consists of following steps. Initially, a new signal is generated from a set of permutations using the axes of the raw signal. Then, a Fourier transform is applied to this new signal, producing the input to the ConvNet. Even with interesting results, the method proposed by Jiang and Yin~\cite{Jiang:2015} presents a remarkable drawback, the high computational cost since their method increases the input matrix size exponentially. This fact prevented us from applying their method in the dataset PAMAP2, as well as from conducting some of our experiments (more details in Section~\ref{sec::experiments}).

Following the hypothesis that different sensor modalities (i.e., accelerometer, gyroscope and magnetometer), should be convolved separately, Ha et al.~\cite{Ha:2015} introduced a zero-padding between each heterogeneous modality to prevent them from being merged during the convolution process. Their architecture consists of two convolutional layers with $32$ and $64$ filters of $3\times3$, respectively. However, due to this architecture, the heterogeneous modalities are convolved together at the second layer. To address this problem, Ha and Choi~\cite{Ha:2016} suggested to introduced a zero-padding before starting the second layer of convolution. For this purpose, a zero-padding was inserted in the feature map generated by the first convolution layer so that the different modalities can be kept separated.

Ha and Choi~\cite{Ha:2016} also demonstrated that ConvNets (2D convolutions) present better results compared with 1D convolutions (Conv1D), RNN or Long Short Term Memory networks (LSTM). In particular, even though recurrent-based networks have been successfully applied to speech recognition~\cite{Graves:2013,Greff:2017} and natural language processing~\cite{Wen:2015, Wang:2016}, there exist few successful works that explore LSTMs in the context of human activity recognition based on wearable sensors~\cite{Yao:2016, Murad:2017}. In general, recurrent-based networks have many hyper-parameters to be set and do no present expressive results when compared to Conv2D. Thus, we do not contemplate this class of approaches in our experiments.
\begin{table*}[!htb]
\centering
\small
\caption{Samples generation processes, metrics and validation protocol employed by different works. While it is clear that accuracy should be used as the metric, there is no sense in terms of validation protocol. \emph{Unknown} denotes that the original paper does not report the technique employed.}
\label{tab::protocols}
\begin{tabular}{|c|c|c|c|}
\hline
Work                                                               & \begin{tabular}[c]{@{}c@{}}Data  Generation\end{tabular} & \begin{tabular}[c]{@{}c@{}}Evaluation Metrics\end{tabular} & \begin{tabular}[c]{@{}c@{}}Validation Protocol\end{tabular} \\ \hline
Pirttikangas~\cite{Pirttikangas:2006} & Semi-Non-Overlapping-Window                                                       & Accuracy                                                          & 4-fold  cross validation                                                    \\ \hline
Suutala et al.~\cite{Suutala:2007}           & Semi-Non-Overlapping-Window                                                       & Accuracy, Precision, Recall                                       & 4-fold  cross validation                                                    \\ \hline
Kwapisz et al.~\cite{Kwapisz:2010}           & Unknown                                                    & Accuracy                                                          & 10-fold  cross validation                                                   \\ \hline
Catal et al.~\cite{Catal:2015}               & Unknown                                                    & Accuracy, AUC, F-Measure                                          & 10-fold cross validation                                                    \\ \hline
Kim et al.~\cite{Kim:2012}                   & Semi-Non-Overlapping-Window                                                       & F-measure                                                     & Unknown                                                        \\ \hline
Kim and Choi~\cite{Kim:2013}                 & Semi-Non-Overlapping-Window                                                       & Accuracy, F-measure                                               & Unknown                                                        \\ \hline
Chen and Xue~\cite{Chen:2015}                & Semi-Non-Overlapping-Window                                                       & Accuracy                                                          & Holdout                                                        \\ \hline
Jiang and Yin~\cite{Jiang:2015}              & Unknown                                                    & Accuracy                                                          & Unknown                                                        \\ \hline
Ha et al.~\cite{Ha:2015}                     & Semi-Non-Overlapping-Window                                                       & Accuracy                                                          & Hold out                                                        \\ \hline
Ha and Choi~\cite{Ha:2016}                   & Semi-Non-Overlapping-Window                                                       & Accuracy                                                          & Leave-One-Subjet-Out                                                           \\ \hline
Yao et al.~\cite{Yao:2016}                   & Semi-Non-Overlapping-Window                                                       & Accuracy.                                                          & Leave-One-Subjet-Out                                                           \\ \hline
Pan et al.~\cite{Panwar:2017}                & Semi-Non-Overlapping-Window                                                       & Accuracy                                                          & Cross validation and Leave-One-Subjet-Out                                                    \\ \hline
Yang et al.~\cite{Yang:2015}                 & Unknown                                                    & Accuracy, F-Measure                                               & Hold out and Leave-One-Subjet-Out                                              \\ \hline
\end{tabular}
\end{table*}
\section{Evaluation Methodology}\label{sec::methodology}
The major concern of the research on human activity recognition based on wearable sensors is the lack of standard protocols to conduct experiments and report results. In other words, simple questions such as \emph{``What is the evaluation metric to report the results?''}, \emph{``How to generate the data samples from the raw signal?''} and \emph{``What are the challenging datasets?"} have not been properly addressed in the existing works.
The disregard of such questions prevents us from comparing the existing works and, as a consequence, it is not possible to determine the {state-of-the-art} in this task.
For instance, while some works use \emph{F-measure} to report the final performance of their methods, others employ accuracy. The problem becomes worse when the authors choose different validation protocols. Moreover, an inadequate process to generate the data samples can bias the real performance of the methods (as it will be shown).

It is important to note that in other application areas, such as object detection~\cite{VOC:2007}, image classification~\cite{ImageNet:2015} and face verification~\cite{LFW:2007, YTF:2011}, the aforementioned issues have been well-defined, which attracts more research efforts towards the application. Therefore, a standardized evaluation is an essential requirement for research in human activity recognition based on wearable sensor data.

Given this overview regarding the problems in wearable sensor data applied to activity recognition, the remaining of this section defines the evaluation methodology of this task.
We start by describing the evaluation metrics and protocols employed by previous works to report the activity recognition performance in the context of wearable sensor data. 
Then, we discuss the traditional sample generation process and introduce the proposed approaches to perform this process.
\subsection{Evaluation Metrics}\label{sec::metrics}
There exists a set of metrics to measure the activity recognition performance, such as accuracy, recall, F-measure and Area Under the Curve (AUC).  Table~\ref{tab::protocols} summarizes the main evaluation metrics employed in human activity recognition based on wearable sensor data.
Among the metrics listed in Table~\ref{tab::protocols}, accuracy and F-measure are obvious choices. In particular, F-measure is more suitable since it is computed using the precision and recall, thereby, it is able to evaluate the activity recognition taking into account two different metrics.
\subsection{Validation Protocols}\label{sec::protocol}
An important step in recognition tasks is to separate the available data into training and testing sets. For this purpose, in the context of human activity recognition based on wearable sensors data, many works apply techniques such as \emph{k}-fold cross validation, leave-one-subject-out, hold-out and leave-one-sample-out (a.k.a leave-one-out). 
The techniques of k-fold cross validation (with $k=10$) and leave-one-subject-out are the traditional preferences, while few works employ the hold-out and leave-one-sample-out (this one due to the large number of executions), as showed in Table~\ref{tab::protocols}.

We highlight that the leave-one-subject-out can be comprehended as a special case of the cross validation, where a subject can be seen as a fold, hence, the number of subjects determine the number of folds. Furthermore, it reflects a realistic scenario where a model is trained in an offline way~\cite{Shoaib:2015} using the samples of some subjects and is tested with samples of unseen subjects. However, by using this protocol, the methods present high variance in accuracy from one subject to another since the same activity can be performed in different ways by the subjects.
\subsection{Sample Generation Process}\label{sec::samples_generation}
The first step to perform human activity recognition based on wearable sensor data is to generate the samples from the raw signal. This process consists of splitting a raw signal into small windows of the same size, referred to as \emph{temporal windows}. Then, the temporal windows from the signals are used as data samples, where they are split into training and test to learn and evaluate a model, respectively.

This section explains the process employed by previous works to generate the temporal windows, \emph{Semi-Non-Overlapping-Window}. It also introduces two novel processes: \emph{Full-Non-Overlapping-Window} and \emph{Leave-One-Trial-Out}, both focuses on addressing the drawback of the existing process.

\subsubsection{Semi-Non-Overlapping-Window} 
This is the most employed process to yield samples (temporal windows) for the activity recognition and it works as follows. Initially, the temporal sliding window technique (defined in Section~\ref{sec::exp_setup}) is applied on the raw signals,  generating a set of data samples. Then, from these data samples, sets for training and test are created using some validation protocol (e.g., 10-fold cross validation).
Since this process considers an overlap of $50\%$ between windows, we called it of \emph{Semi-Non-Overlapping-Window}.

A notable drawback of this process is that it is highly biased. This occurs because a window $i$ and $i{+1}$ can appear in different folds of the cross validation (or any other protocol). Thereby, $50\%$ of the content of these windows are equal because they present overlapping. As a consequence, $50\%$ of a sample can appear in both training and testing at the same time, biasing the results. In other words,  training and testing samples might be very close temporally, generating skewed results. Therefore, based on the second column of Table~\ref{tab::protocols}, the results reported by the previous works do not reflect their real performance.
 
We emphasize that, in work work, the term \emph{bias} refers to the fact that part of the {sample's content} appears in the training and testing, simultaneously. According to the experiments, the methods drop the accuracy notably when changing from this process to another without this bias.

It is important to mention that on the leave-one-subject-out validation protocol, the semi-non-overlapping-window technique is not affected by bias, which is desirable, since the samples of training and testing are separated by subjects. Therefore, the raw signal used to yield the samples (which can be temporally close) will either appear in the training or in the testing, but not in both.

\subsubsection{Full-Non-Overlapping-Window} 
A simple way to handle the bias problem of the aforementioned process is to ensure that the windows have no temporal overlapping, guaranteeing that part of the window's content does not appear in the training and testing, simultaneously.
For this purpose, we propose the use of non-overlapping windows (overlap equal to zero between the temporal windows) to generate samples, process referred to as \emph{Full-Non-Overlapping-Window}.

Even though the proposed full-non-overlapping-window process prevents the bias, it is has the disadvantage of providing a reduced number of samples when compared to the semi-non-overlapping-window process (around $1.10$ times fewer samples\footnote{Value computed using the average of all the datasets.}) since the temporal windows no longer overlap.
\subsubsection{Leave-One-Trial-Out} As we argued earlier, each process has a drawback that might cause a negative impact on the methods. For instance, semi-non-overlapping-window process produces biased results while the full-non-overlapping-window process generates few samples. 

To avoid the aforementioned problems, we propose the \emph{Leave-One-Trial-Out} process. A trial is the raw signal of one sequence of activities (or a single activity) performed by one subject. In this work, we propose to use the trials to ensure that samples generated by the same signal do not appear in the training and testing, simultaneously. To achieve this goal, we apply 10-fold cross validation on the raw signals, guaranteeing that the same trial (the same raw signal) does not appear in both training and testing. This is possible since k-fold cross validation ensures that the same sample (here, the trial - raw signal) appears in one fold only. Finally, for each fold, we generate the data samples (temporal windows) from raw signals, following the same process as in semi-non-overlapping-window. 
Throughout this process, we do not change the number of folds, this way, the number of folds is defined by k-fold cross validation (where k=10). It is important to mention that it is not possible/adequate to employ the trial as a criterion to generate the folds since a single trial, in most cases, does not contain all the activities.

By using this process, we avoid: (1) bias since part of the window content never appear in the training and testing  at the same time, and (2) small number of samples because the overlapping used in this process is the same employed in the semi-non-overlapping-window.
\subsection{Standardization of the Wearable Sensors Datasets}\label{sec::methodology_datasets}
Nowadays, there are many available datasets to perform human activity recognition based on wearable sensor data. These datasets present a wide range of sampling rate, number of activity categories and available sensors (see Table~\ref{tab::datasets}), which enable us to evaluate the activity recognition in different scenarios. However, the lack of standardization of the captured data renders difficulties to develop a general framework able to perform activity recognition on all the datasets. 

In general, the wearable sensors datasets can be divided into two groups with respect to the manner in which the activities were captured. The first group consists of activities where the user performs all the activities freely, i.e., there is no pause between the execution of one activity and the next one (e.g., MHEALTH, PAMAP2 and WISDM). The second group contains activities captured separately, i.e., a single activity is performed at a time (e.g., USC-HAD, UTD-MHAD and WHARF). This difference between these groups makes hard to perform a unified evaluation. Therefore, it is important to consolidated them into a single form. Intuitively, the first group of datasets can be converted into the second group, while the inverse is not possible, thereby, in this work we standardize all the datasets to simulate the second type of dataset.

As a final note, since the activity recognition datasets involve human participants, the ETHICS approval is required. This approval is of responsibility of the authors that proposed the datasets and can be found in the original works where the datasets were proposed.
\begin{table*}[!ht]
	\centering
	\caption{Main features of the datasets used in this work. The number of samples was computed using the semi-non-overlapping-window process with leave-one-subject-out validation protocol. Acc, Gyro, Mag and Temp indicate accelerometer, gyroscope, magnetometer and temperature, in this order.}
	\label{tab::datasets}
	\begin{tabular}{|c|c|c|c|c|c|c|c|}
		\hline
		Dataset                               & Frenquency (Hz) & \#Sensors                  & \#Activities & \#Subjects & \#Trials & \#Samples & Balanced \\ \hline
		MHEALTH~\cite{Banos:2014}      & $50$            & $3$ (Acc, Gyro, Mag)       & $12$         & $10$       & $262$    & $2555$    & True     \\ \hline
		PAMAP2~\cite{Reiss:2012}                                & $100$           & $4$ (Acc, Gyro, Mag, Temp) & $12$         & $10$       & $108$    & $7522$    & False    \\ \hline
		USCHAD~\cite{Zhang:2012}       & $100$           & $2$ (Acc, Gyro)         & $12$         & $15$       & $840$    & $9824$    & False    \\ \hline
		UTD-1~\cite{ChenChen:2015} & $50$            & $2$ (Acc, Gyro)            & $21$         & $9$        & $617$    & $3771$    & True     \\ \hline
		UTD-2~\cite{ChenChen:2015} & $50$            & $2$ (Acc, Gyro)            & $5$          & $9$        & $190$    & $1137$    & True     \\ \hline
		WHARF~\cite{Bruno:2015}        & $32$            & $1$ (Acc)                  & $12$         & $17$       & $884$    & $3871$    & False    \\ \hline
		WISDM~\cite{Lockhart:2011}     & $20$            & $1$ (Acc)                  & $7$          & $36$       & $402$    & $20846$   & False    \\ \hline
	\end{tabular}
\end{table*}
\section{Experimental Evaluation}\label{sec::experiments}
We start this section by describing the experimental setup and the datasets in Section~\ref{sec::exp_setup} and ~\ref{sec::datasets}, respectively. Afterwards, we present the experiments demonstrating the influence of the sample generation process on the activity recognition performance (Section~\ref{exp::comparison_protocols}). 
Then, we evaluate the impact of using subjects
to separate the training and testing samples (Section~\ref{exp::losoTen}), and investigate the activity recognition performance according to the datasets employed (Section~\ref{exp::comparison_datasets}). Finally, we compare the previous works using statistical tests in Section~\ref{exp::statistical_evaluation}, and discuss their results in  Section~\ref{sec::state_of_the_art}, where we define the {state-of-the-art} in activity recognition based on wearable data.

According to Table~\ref{tab::protocols}, accuracy is the evaluation metric most employed by existing works, therefore, we have selected it to assess the activity recognition quality. To separate the data into training and testing ({validation protocol}), we used the $10$-fold cross validation (since it is a common choice, as seen in Table~\ref{tab::protocols}), except for the leave-one-subject-out protocol, where the folds are defined by the number of subjects.
\subsection{Experimental Setup}\label{sec::exp_setup}
\subsubsection{Temporal Sliding Window}\label{sec::tempora_window}
To increase the number of samples and enable the activity recognition to operate with a small latency (expected for real-time activity recognition), the works in the literature employ the temporal sliding window technique~\cite{Kim:2013,Chen:2015,Ha:2015}. This technique consists of dividing the sample into subparts (windows) and considering each subpart as an entire activity. Specifically, each window becomes itself a sample that will be associated to a class label after its classification. A temporal sliding window can be defined as
\begin{equation}
\label{eq::temporal_sliding_window}
  \begin{split}
     w =  [s_{k-t},..., s_{k-2}, s_{k-1}, s_k ]^\top,\\
  \end{split}
\end{equation}
where $k$ represents the current signal captured by the sensor and $t$ denotes the temporal sliding window size. The windows might overlap and the ones that do not fit within the temporal window are dropped. In other words, windows with the size smaller than $t$ are discarded. Based on previous works~\cite{Morris:2014,Song:2017}, we are using $t$ equals to $5$ seconds, which represents a good trade-off between the number of discarded samples and recognition accuracy.
\subsubsection{Convolutional Neural Network Setup} Different from handcrafted approaches, where the authors provide enough information for reproducibility, most works based on ConvNets omit some important parameters, such as number of epochs, batch size and the optimizer used. To handle this problem and provide a fair comparison among this group of approaches, we set these parameters as follows. 
The maximum number of epochs was set as $200$ and the method stops its training when the loss function reaches a value less or equal to $0.2$. These values were set empirically by observing the trade-off between execution time and accuracy. Similarly, the batch size was set to $1000$, except for the PAMAP2 dataset, where this value was of $250$ due to memory issues. Finally, we employ the Adadelta optimizer~\cite{Zeiler:2012} (except for the methods where the optimizer was specified by the author). In preliminary experiments, this optimizer presented the best accuracy when compared to SGD and RMSprop~\cite{LeCun:2012}, besides providing an efficient execution time.
\begin{figure*}[!ht]
	\centering
	\subfigure[Results using the semi-non-overlapping-window and cross validation (SNCV) combination.] {\includegraphics[width=0.475\linewidth]{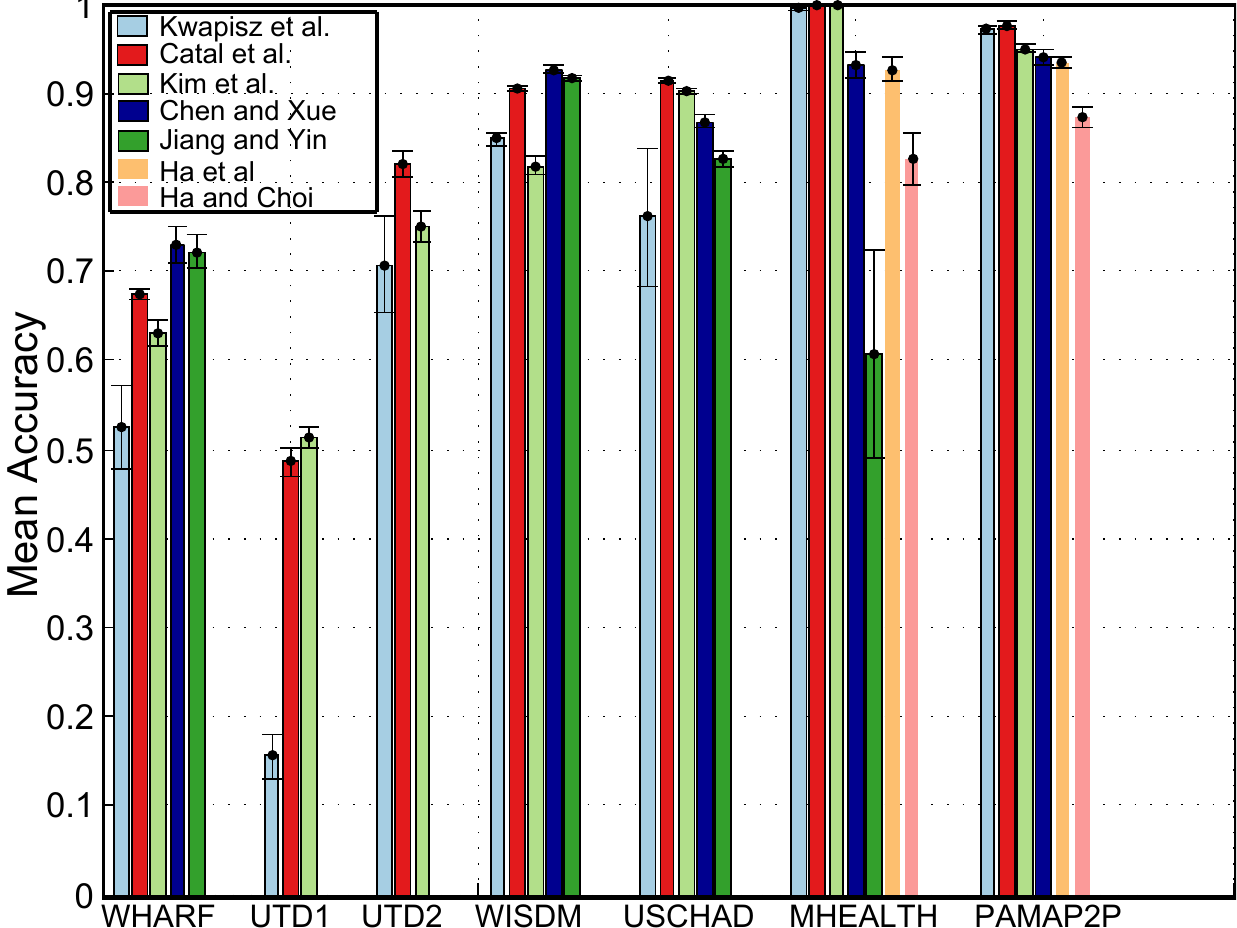}}
	\hspace{4mm} 
	\subfigure[Results using the full-non-ovelapping-window and cross validation (FNCV) combination.] {\includegraphics[width=0.475\linewidth]{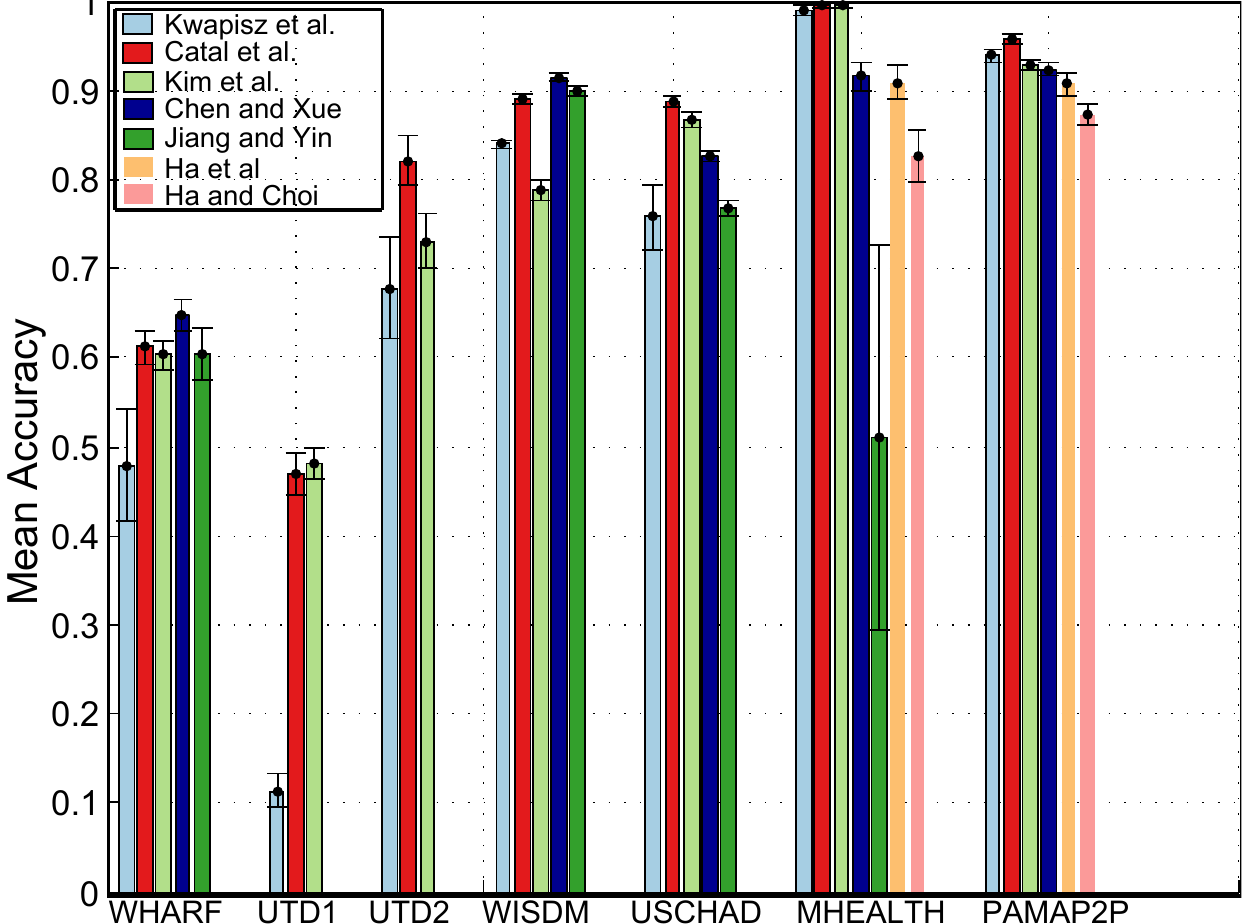}}
	\subfigure[Results using the leave-one-trial-out and cross validation (LTCV) combination] {\includegraphics[width=0.475\linewidth]{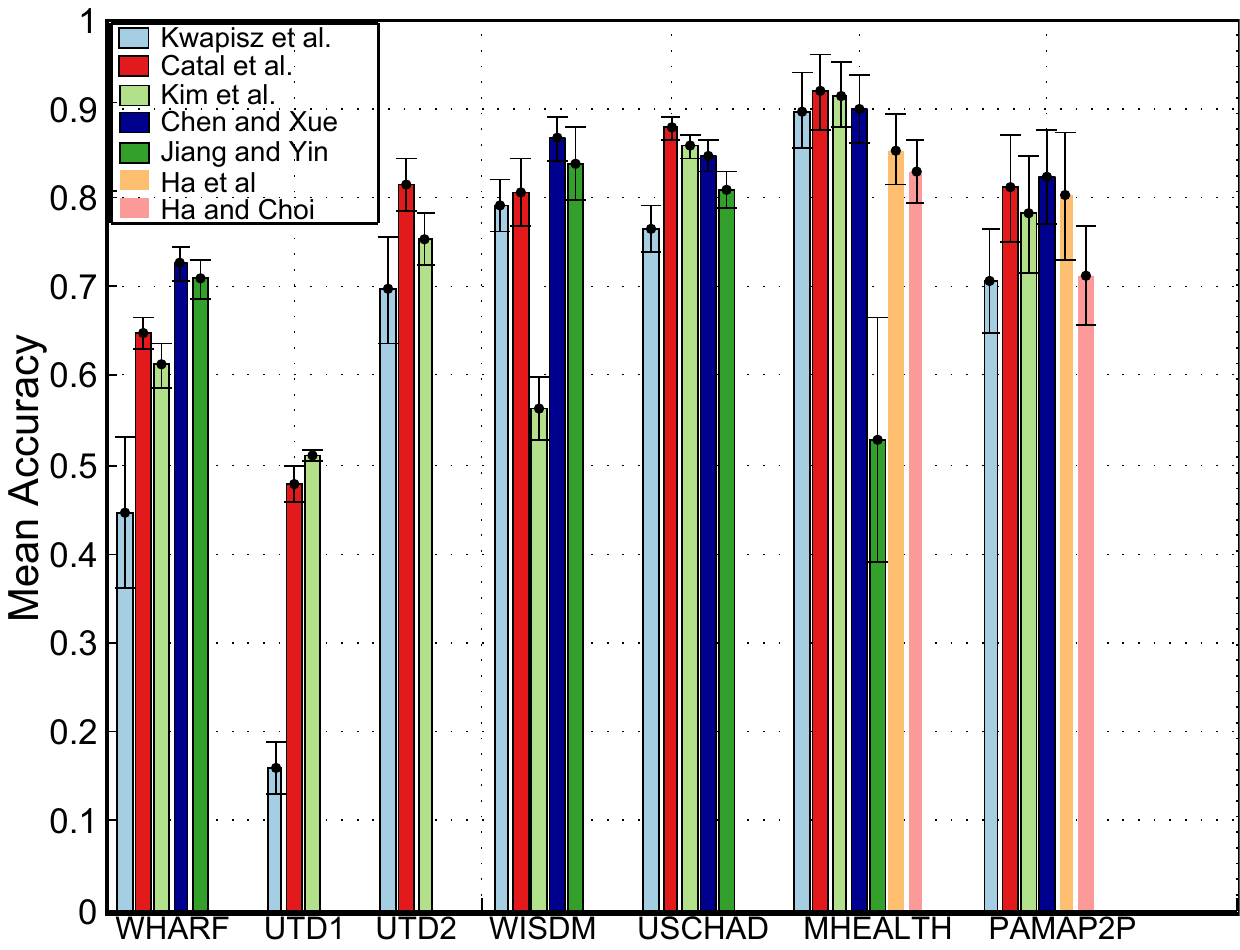}}
	\hspace{4mm} 
	\subfigure[Results using the semi-non-overlapping-window and leave-one-subject-out (SNLS) combination]	
	{\includegraphics[width=0.475\linewidth]{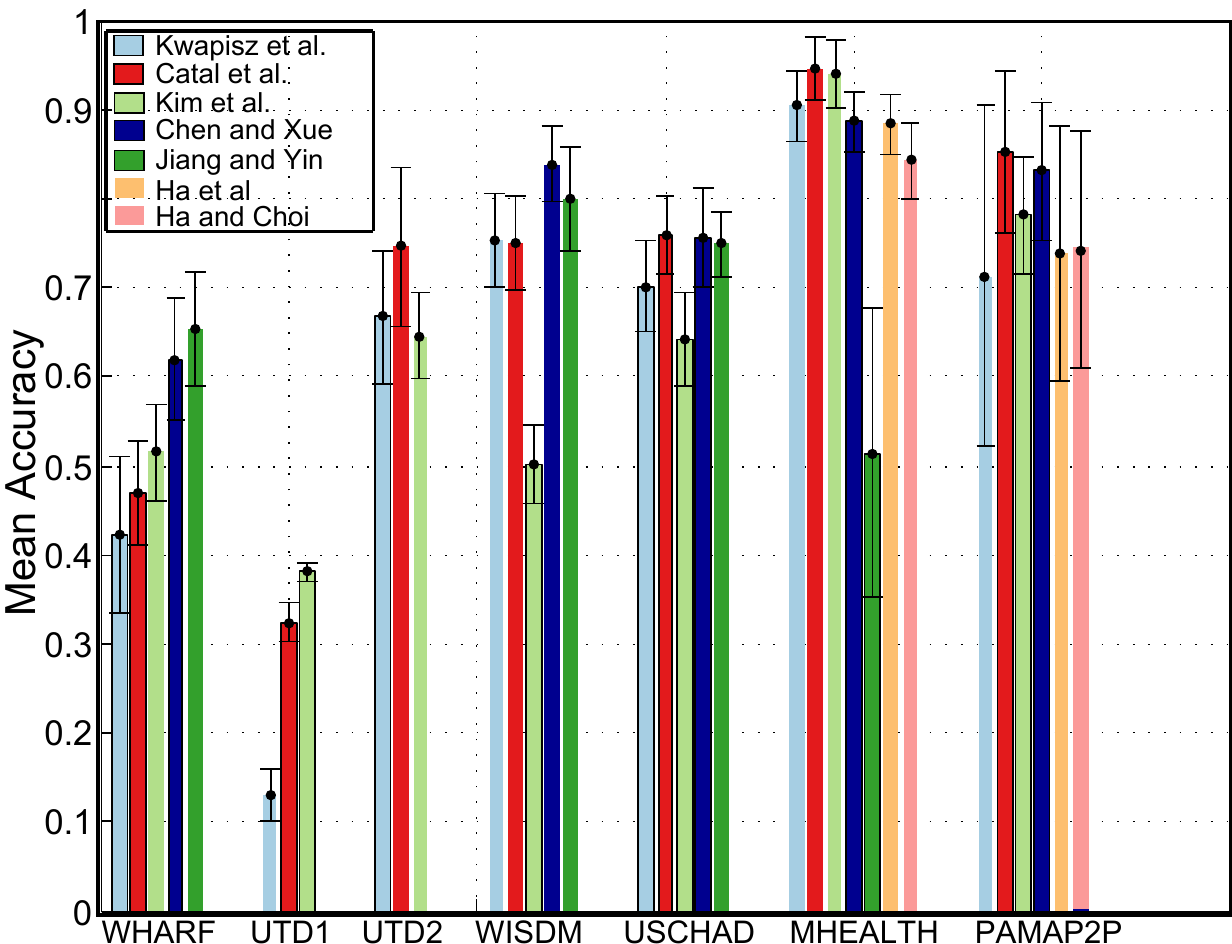}}
	\caption{Mean accuracy achieved by the methods using different process to generate the data samples (best visualized in color). Black bars denote the confidence interval.}
	\label{fig:state_of_the_art}
\end{figure*}

It is important to mention that the employment of deep architectures and large convolutional kernels makes impracticable the use of ConvNets in some datasets where the sampling rate is small, i.e., WHARF and WISDM (see Table~\ref{tab::datasets}). In deep architectures, this occurs since the convolution process produces feature maps smaller than the input presented to it and its size can reach zero in deeper layers of a ConvNet. Therefore, it was not possible to execute some of the ConvNets considered in this work for all datasets.
\subsection{Datasets}\label{sec::datasets}
The datasets evaluated in our study consider a variety of sampling rate, number of activities categories and degree of difficulty. To select these datasets, we consider the ones which provide enough information regarding the capturing of the data, activities, subjects and the employed sensors. Additionally, we label a dataset as unbalanced when its largest class has four times more samples than the smallest class. 

The main features of the datasets are summarized in Table~\ref{tab::datasets} and briefly described as follows. The MHEALTH and PAMAP2 datasets consist of activities captured from sensors placed on the subject's chest, wrist and ankle. Similarly, the activities of the WHARF dataset were obtained from a sensor placed on the subject's wrist.
Different from these datasets, the activities of the WISDM dataset were captured using a sensor located in the user's pocket focusing on convenience and comfort for the subject.
Finally, the activities of the UTD-MHAD dataset were captured using two configurations, the sensor placed at the subject's wrist and subject's thigh, UTD-1 and UTD-2, in this order.
\subsection{Sample Generation Processes and Validation Protocols}\label{exp::comparison_protocols}
\begin{table}[!b]
\centering
\caption{Combinations between samples generation processes and validation protocols. The symbol '-' denotes we do not consider the respective combination.}
\label{tab::combinations}
\small
\begin{tabular}{c|c|c|}
\cline{2-3}
& \begin{tabular}[c]{@{}c@{}}10-Fold\\ Cross Validation\end{tabular} & \begin{tabular}[c]{@{}c@{}}Leave-One\\Subject-Out\end{tabular} \\ \hline
\multicolumn{1}{|c|}{\begin{tabular}[c]{@{}c@{}}Semi-Non\\ Overlapping-Window\end{tabular}} &  SNCV & SNLS \\ \hline
\multicolumn{1}{|c|}{\begin{tabular}[c]{@{}c@{}}Full-Non\\ Overlapping-Window\end{tabular}} &   FNCV  &    -  \\ \hline
\multicolumn{1}{|c|}{\begin{tabular}[c]{@{}c@{}}Leave-One\\ Trial-Out\end{tabular}}         &     LTCV    &  -  \\ \hline
\end{tabular}
\end{table}
This experiment intends to demonstrate that there is a considerable variance in the results achieved by the methods when different sample generation processes and validation protocols are considered. In particular, we can use different combinations of sample generation processes and validation protocols to perform activity recognition, as seen in Table~\ref{tab::combinations}. However, to reduce the number of experiments, we use the proposed sample generation processes on the 10-fold cross validation only, which is the most employed validation protocol, reducing the number of possible combinations to four. In addition, once selected a combination of Table~\ref{tab::combinations}, we ensure that the training and testing samples are the same to all the methods. In this way, we provide an adequate and fair comparison.

Figure~\ref{fig:state_of_the_art} shows the mean accuracy and the confidence interval of the methods when evaluated on different combinations of sample generation processes and validation protocols (see Table~\ref{tab::combinations}). According to Figure~\ref{fig:state_of_the_art}(a), it is possible to note that the combination SNCV reports the highest mean accuracy when compared to the other combinations. 
This happens due to bias produced by its sample generation process, where the content of a window can appear in both training and testing. Therefore, the works that employ the SNCV combination might have their results highly skewed.

Figure~\ref{fig:state_of_the_art}(b) shows the results of the methods using the FNCV combination. According to the results, the SNCV and FNCV achieve similar mean accuracy, where the second one presents, slightly, inferior results. This effect is expected since FNCV also has a sampling bias because the trial used to generate the samples (which do not present overlapping) can produce samples of training and testing at the same time. In other words, parts of the same trial can appear in training and testing.
On the other hand, when the LTCV combination is considered (Figure~\ref{fig:state_of_the_art}(c)), the mean accuracy drops significantly. This is a consequence of the data generation, which do not have any type of bias. It is possible to note this behavior by observing the results achieved in the PAMAP2 and MHEALTH datasets, where the methods had its accuracy reduced drastically when compared to the results of the SNCV and FNCV combinations. 

Figure~\ref{fig:state_of_the_art}(d) illustrates the results achieved on the SNLS combination, where the mean accuracy had the smallest performance. This occurs since SNLS is invariant to bias, since the samples of training and testing are separated by subjects, which means the raw signal used to yield the samples either will appear in the training or testing only, causing an effect similar to the proposed leave-one-trial-out.

Based on the aforementioned discussion, it is possible to note that there exists a high variance in the results depending on the combinations between samples generation processes and validation protocols. To demonstrate this, let us compare the number of methods which achieved a mean accuracy above $80\%$\footnote{We select this value empirically just for discussion.} to a determinate data sample generation process, by considering all the datasets. From SNCV to FNCV (Figure~\ref{fig:state_of_the_art}(a) and (b)) and FNCV to LTCV (Figure~\ref{fig:state_of_the_art}(b) and (c)), this number decreased from $21$ to $20$ and from $20$ to $16$, in this order. These values indicate that the activity recognition becomes more difficult with respect to the samples generation process employed. In particular, this variance in the results is a direct effect of the bias introduced during the process of generating the data samples, where part of the window's content appear both in training and testing.
This remark is easier to notice when we compare SNCV and SNLS, Figures~\ref{fig:state_of_the_art}(a) and (d). From this comparison, the number of methods which achieved an accuracy above $80\%$ decreased from $21$ to $10$, when using SNLS instead of SNCV.

\begin{figure}[!t]
	\centering
	\subfigure[SNCV] {\includegraphics[width=0.325\linewidth]{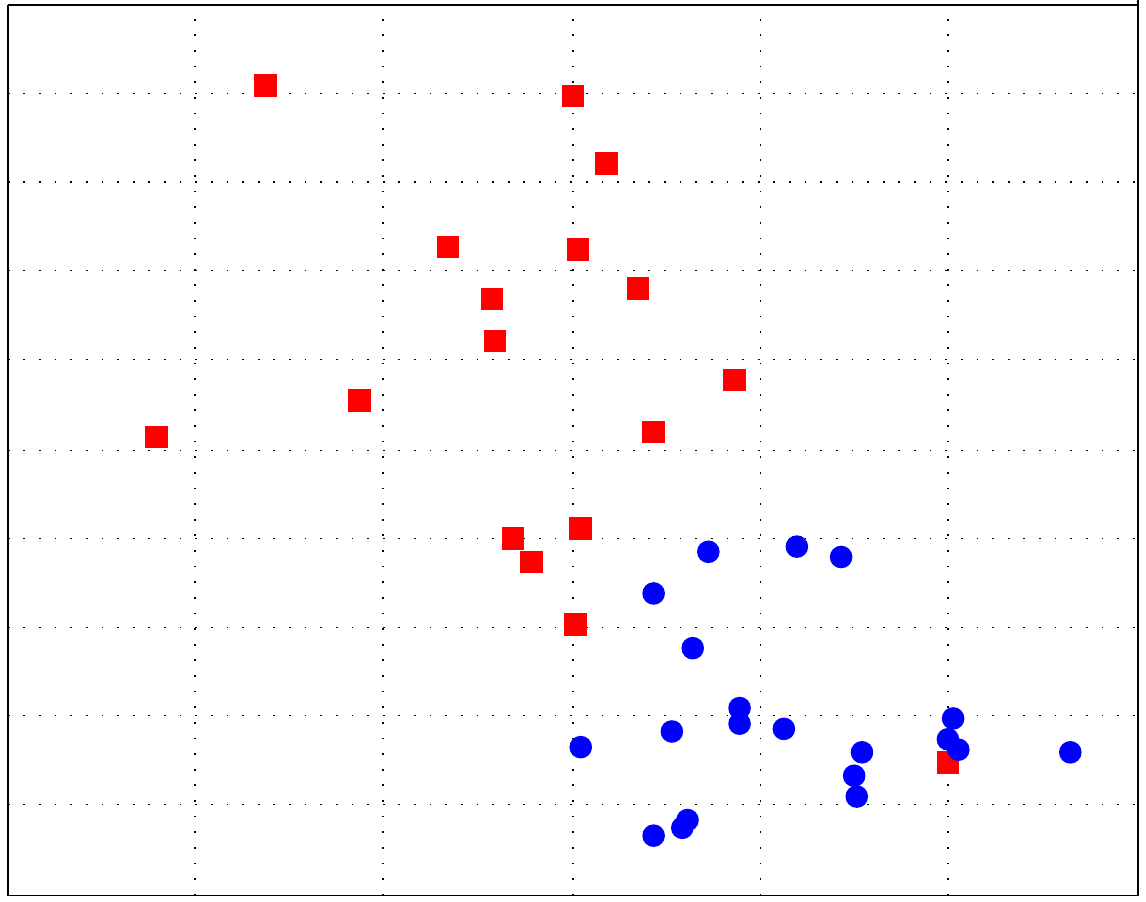}} 
	\subfigure[FNCV] {\includegraphics[width=0.325\linewidth]{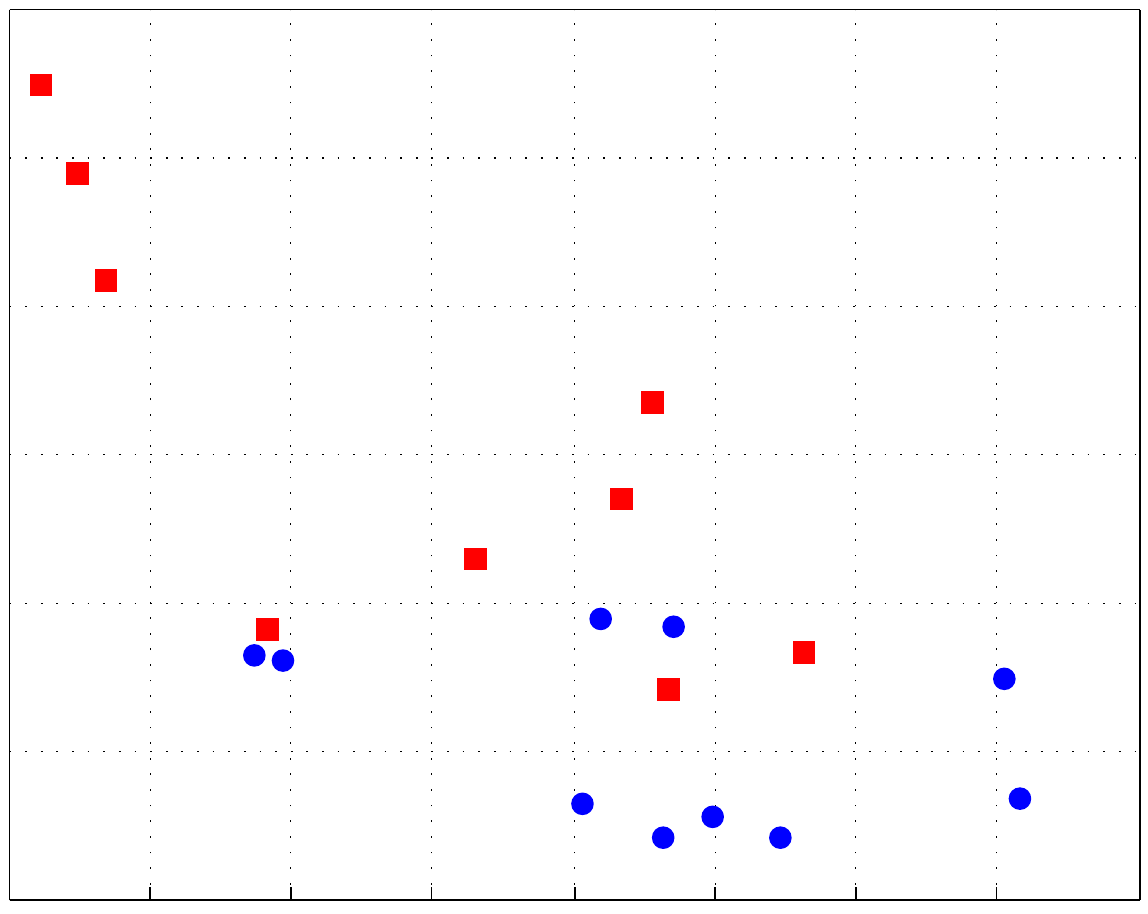}}
	\subfigure[LTCV] {\includegraphics[width=0.325\linewidth]{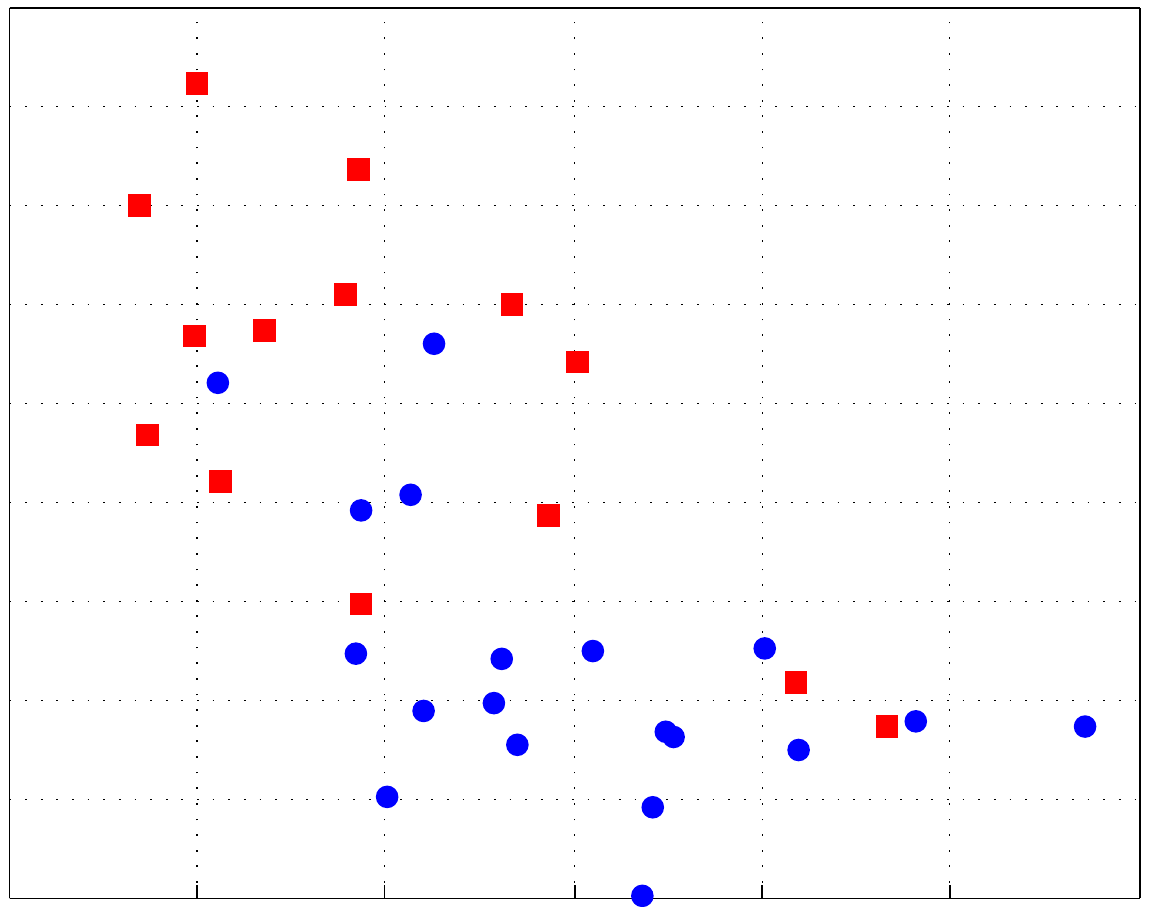}}
	\caption{Feature space provided by the two first components of Linear Discriminant Analysis on the features suggested by~\cite{Kwapisz:2010}. Note that from Figure (a) to (c) the separability of the classes is decreasing, which is a consequence of the process used to generate the data samples. To provide a better visualization, we show only two activities categories (best visualized in color).}
	\label{fig:lda}
\end{figure}

To illustrate the data behavior when the samples generation process is changed, we set the validation protocol (10-fold cross validation) and vary the samples generation process, then, we project the training samples onto the two first components of Linear Discriminant Analysis (LDA)~\cite{Markopoulos:2017}, Figure~\ref{fig:lda}. From this figure, it is clear that the class separability decreases according to the process employed. 

\begin{figure}[!t]
\centering
\includegraphics[scale=0.5]{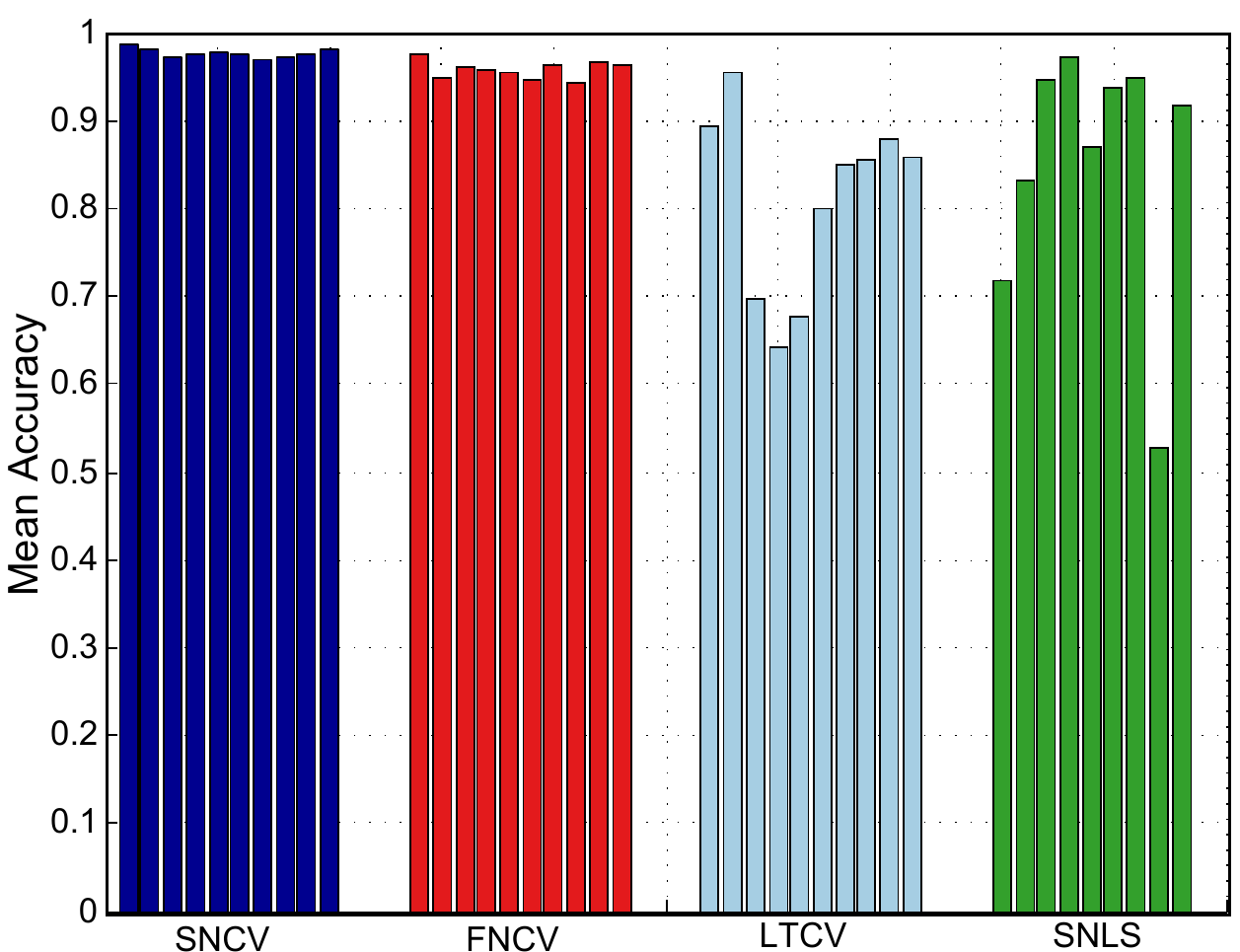}
\caption{Accuracy obtained by each fold using different combinations of validation protocols and samples generation processes (best visualized in color).}
\label{fig:folds}
\end{figure}

Another point to be evaluated regarding the combinations in Table~\ref{tab::combinations} is the accuracy variance between the folds of the validation protocol. To that, we select the method of Catal et al.~\cite{Catal:2015} and report their accuracy obtained for each fold using the PAMAP2 dataset\footnote{Specifically, we choose this method since it presents the best performance and we select the PAMAP2 dataset because it provides exactly $10$ subjects, which enables us to compare the i-\emph{th} subject with the i-\emph{th} fold.}, Figure~\ref{fig:folds}. Figure~\ref{fig:folds} shows that SNCV and FNCV present minimal variance. However, it increases when the methods are evaluated on the LTCV and SNLS combinations.

Based on the experiments conducted in this section, we demonstrated that the methods have their performance extremely associated with the combination between samples generation process and validation protocol. For instance, semi-non-overlapping-window using cross validation introduces bias, which is inadequate for conducting experiments, while semi-non-overlapping-window using leave-one-subject-out is bias-invariant. Unfortunately, according to Table~\ref{tab::protocols}, the majority of the methods employ the first combination. On the other hand, we showed that this bias can be slightly reduced or completely removed with the employment of our proposed sample generation process.
\subsection{Limitations of Leave-One-Subject-Out}\label{exp::losoTen}
In this experiment, we intend to show the drawback of leave-one-subject-out validation protocol and propose an alternative.
According to Figure~\ref{fig:state_of_the_art}(d) and Table~\ref{tab::loso}, the confidence interval (which is directly affected by the accuracy variance of the folds) is extremely high, as indicated in datasets such as WHARF, MHEALTH and PAMPA2P. This behavior is due to high variability in the samples provided by the folds (recall that for leave-one-subject-out, folds are the subjects). Specifically, different subjects can perform the activities in distinct ways; hence, the learned model is not able to produce enough generalization to correctly classifying the testing subject.
A drawback regarding this issue is that the methods become statistically equivalent ({since the confidence intervals are large and overlap each other}~\cite{Raj:1990}). 

To face this problem, we propose SNLS$\times10$. The idea behind SNLS$\times10$ is to measure the accuracy and the confidence interval by using the variability from the training samples, which is small, instead of the variability from the subjects, which is high causing a large confidence interval. With this purpose, we execute the leave-one-subject-out $10$ times and for each turn we select just $80\%$ of the training samples (the same samples used in traditional SNLS) to learn the model.

\begin{figure}[!t]
\centering
\includegraphics[scale=0.5]{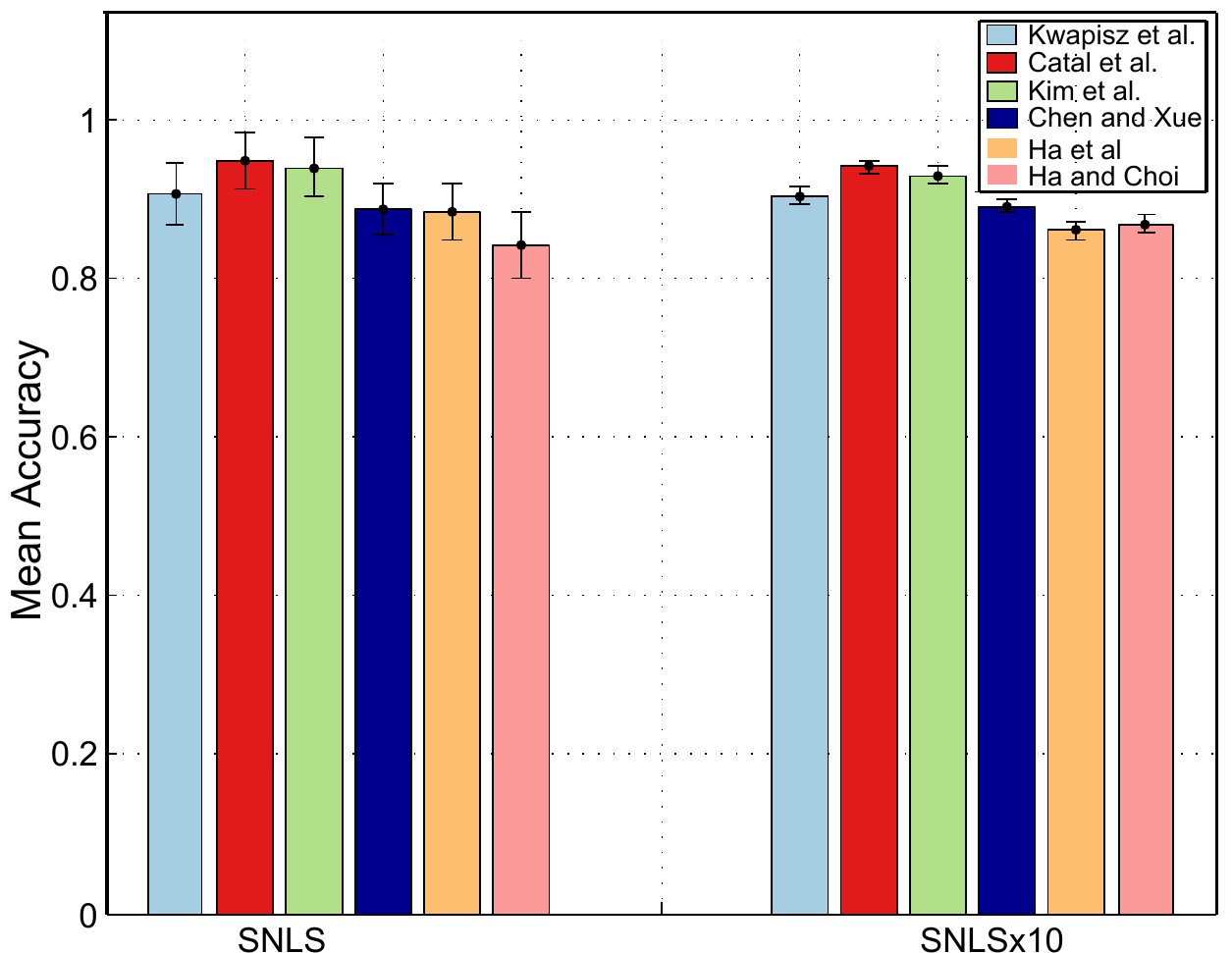}
\caption{Comparison of the confidence interval between SNLS and SNLS$\times10$ (best visualized in color).}
\label{fig:LOSOx10}
\end{figure}

Figure~\ref{fig:LOSOx10} shows that performing SNLS$\times10$ decreases the confidence interval when compared to the traditional SNLS, enabling us to compare the methods using the confidence interval. According to  Figure~\ref{fig:LOSOx10}, methods of Catal et al.~\cite{Catal:2015} and Kim et al.~\cite{Kim:2012} achieved the Top 1 best accuracy besides being statistically different from the remaining methods. The Top 2 best accuracy was obtained by the methods of Kwapisz et al.~\cite{Kwapisz:2010} and  Chen and Xue~\cite{Chen:2015}, which are statistically equivalent. Finally, the Top 3 best result was achieved by Ha et al.~\cite{Ha:2015} and Ha and Choi~\cite{Ha:2016}. 

Note that it is not possible to list the rank above using the traditional SNLS combination since all the methods (except~\cite{Ha:2016}) are statistically equivalent. However, the computational cost of SNLS$\times10$ increases because it is necessary to execute more turns of the leave-one-subject-out  ($\times10$). This fact forced us to perform the SNLS$\times10$ only in the MHEALTH dataset where it is possible to execute all the methods.\footnote{Due to the high number of executions and memory constraints, in this experiment was not possible to execute the method of Jiang and Yin~\cite{Jiang:2015}.}

According to the results achieved in this experiment, we conclude that, though SNLS is an adequate combination to avoid bias and to force the model to be generalized, it is unsuitable to perform statistical tests among the methods.

\subsection{Comparison of Datasets}\label{exp::comparison_datasets}
This experiment intends to determine which are the most challenging datasets in human activity recognition based on wearable sensor data. As we argued in previous experiments, semi-non-overlapping-window and full-non-overlapping-window when combined with cross validation, are not suitable to conduct experiments. Therefore, in this experiment, our discussion is conducted considering the combinations of leave-one-trial-out with cross validation and semi-non-overlapping-window with leave-one-subject-out (LTCV and SNLS, in this order, as defined in Table~\ref{tab::combinations}).

In this experiment, we measure the mean accuracy obtained by all the methods on each dataset (last row in Tables~\ref{tab::loto} and \ref{tab::loso}). According to the results, the most challenging dataset is UTD-1, with a mean accuracy of $38.25\%$ and $27.84\%$ when considering the LTCV and SNLS combinations, respectively. 
We believe that the low accuracy is due to two main reasons: the low sampling rate and the large number of activities. In particular, the large number of activities to be recognized in UTD-1 makes more challenging the recognition as can be observed by the contrast in accuracy regarding UTD-2, where there are a smaller number of activities. Similarly, on the WHARF dataset, the methods also achieved a low performance, with a mean accuracy of $62.76\%$ and $53.55\%$ when employing the LTCV and SNLS combinations, respectively.
On the other hand, the methods achieved the best performance on the MHEALTH dataset, where the mean accuracy was superior to $83\%$. This is an effect of the high sampling rate and the number of available sensors provided by this dataset, which make easier the recognition of the activities.

Observe that our discussion above considers the mean accuracy of all methods, which can be skewed by single methods, for instance, the methods of Kim et al.~\cite{Kim:2012} and Jiang and Yin~\cite{Jiang:2015} on the WISDM and MHEALTH datasets, respectively. However, by examining the datasets using only the best method, these claims still remain valid.
\subsection{Statistical Evaluation}\label{exp::statistical_evaluation}
This experiment intends to show whether the performance achieved by works in human activity recognition is, de-facto, statistically superior or equivalent to other. In addition, it focuses on showing that the proposed leave-one-trial-out with cross validation (LTCV) is more suitable than the combination semi-non-overlapping-window with leave-one-subject-out (SNLS) for conducting statistical tests.

Following Jain~\cite{Raj:1990}, we perform a \emph{unpaired t-test}, which works as follows. For each pair of methods, we computed the confidence interval, using a confidence of $90\%$, from the difference between their mean accuracies.
In the cases where the resulting confidence interval includes the zero value, the methods are statistically equivalent. Otherwise, they are statistically different. Note that this process is the same suggested in~\cite{Raj:1990}.
Due to the high number of comparisons, we perform this evaluation on MHEALTH and PAMAP2 datasets only, which are the ones where it is possible to execute a larger number of works.
\begin{table}[!t]
\centering
\caption{Number of times where a method was statistically equivalent to other.}
\label{tab::paired_test}
\begin{tabular}{|c|c|c|c|c|}
\hline
               & \multicolumn{2}{c|}{MHEALTH} & \multicolumn{2}{c|}{PAMAP2} \\ \hline
               & LTCV          & SNLS         & LTCV         & SNLS         \\ \hline
Kwapisz et al.~\cite{Kwapisz:2010} & $4$             & $4$            & $2$            & $5$            \\ \hline
Catal et al.~\cite{Catal:2015}   & $3$             & $2$            & $3$            & $5$            \\ \hline
Kim et al.~\cite{Kim:2012}     & $3$             & $2$            & $5$            & $5$            \\ \hline
Chen and Xue~\cite{Chen:2015}   & $4$             & $3$            & $3$            & $5$            \\ \hline
Jiang and Yin~\cite{Jiang:2015}  & $0$             & $0$            & $0$            & $0$            \\ \hline
Ha et al.~\cite{Ha:2015}      & $3$             & $3$            & $3$            & $5$            \\ \hline
Ha and Choi~\cite{Ha:2016}    & $1$             & $1$           &  $2$          &   $5$            \\ \hline
\end{tabular}
\end{table}
Table~\ref{tab::paired_test} shows the number of times where a method was statistically equivalent to another using the unpaired t-test. 

For both datasets and combinations evaluated, the method of Jiang and Yin~\cite{Jiang:2015} is the one with the lower number of statistical equivalence. However, sometimes, their accuracy was statistically inferior. On the other hand, on the PAMAP2 dataset and the SNLS combination, the methods of Ha et al.~\cite{Ha:2015} and, Ha and Choi~\cite{Ha:2016} were the ones with the smaller number of statistical equivalence. 

According to Table~\ref{tab::paired_test}, a considerable amount of methods shown to be statistically equivalents. This is an effect of the large confidence interval (denoted by brackets in Tables~\ref{tab::snow}-\ref{tab::loso} and black bars in Figure~\ref{fig:state_of_the_art}), caused by the high variance in accuracy, as seen in Figure~\ref{fig:folds}. In particular, the LTCV combination has a smaller variance than SNLS ($0.0107$ against $0.1448$, respectively). As a consequence, LTCV presented a smaller number of methods which are statistically equivalents (mainly on the PAMAP2 dataset). This evidence shows that our leave-one-trial-out when using cross validation is adequate to conduct statistical tests.
\subsection{The State-of-The-Art}\label{sec::state_of_the_art}
Our last experiment focuses on defining the state-of-the-art methods in human activity recognition based on wearable sensor data. To this end, following previous experiments, we discuss the results using the combination of LTCV and SNLS. In addition, since most of the methods are statistically equivalent, as shown in the earlier experiment, we are not considering statistically difference to determinate if a method is superior than another.

Based on the results presented in Tables~\ref{tab::loto} and \ref{tab::loso}, we report the number of datasets where the methods achieved the best (Top 1) accuracy. Additionally, since there exists a high variance in accuracy among datasets, as seen in Tables~\ref{tab::loto}-\ref{tab::loso} and Figure~\ref{fig:state_of_the_art}, we consider the second (Top 2) and the third (Top 3) best accuracy, aiming to make a fairer comparison. Table~\ref{tab::state_of_the_art} summarizes these results.
\begin{table}[!b]
\centering
\caption{Number of datasets where a method achieved the first (Top1), second (Top2) and third (Top3) best performance. Values after '/' denote the number of datasets where were possible to execute the method.}
\label{tab::state_of_the_art}
\begin{tabular}{|c|c|c|c|c|c|c|}
\hline
                                          & \multicolumn{3}{c|}{LTCV} & \multicolumn{3}{c|}{SNLS} \\ \hline
Method                                    & Top1       & Top2      & Top3      & Top1       & Top2      & Top3      \\ \hline
Kwapisz et al.~\cite{Kwapisz:2010} &$0/7$ &$0/7$&$1/7$& $0/7$& $2/7$ & $0/7$ \\ \hline
Catal et al.~\cite{Catal:2015}     &$3/7$ & $2/7$ & $2/7$ &  $4/7$ & $1/7$ & $0/7$      \\ \hline
Kim et al.~\cite{Kim:2012}         & $1/7$ & $3/7$ & $1/7$ & $1/7$ & $1/7$ & $3/7$ \\ \hline
Chen and Xue~\cite{Chen:2015}  & $3/5$ & $0/5$ & $2/5$ & $1/5$ & $3/5$ & $0/5$ \\ \hline
Jiang and Yin~\cite{Jiang:2015}    & $0/4$ & $2/4$ & $0/4$ & $1/4$ & $1/4$ & $1/4$ \\ \hline
Ha et al.~\cite{Ha:2015}           & $0/2$ & $0/2$ & $1/2$ &  $0/2$ & $0/2$ & $0/2$ \\ \hline
Ha and Choi~\cite{Ha:2016}         & $0/2$ & $0/2$ & $0/2$ & $0/2$ & $0/2$ & $0/2$  \\ \hline
\end{tabular}
\end{table}
\begin{table*}[!htb]
\centering
\tiny
\caption{Mean accuracy and confidence interval using the Semi-Non-Overlapping-Window and Cross Validation (SNCV) combination. The symbol '-' denotes which was not possible to execute the method on the respective dataset.}
\label{tab::snow}
\begin{tabular}{|c|c|c|c|c|c|c|c|c|}
\hline
& MHEALTH & PAMAP2 & USCHAD & UTD-1 & UTD-2 & WHARF & WISDM & Mean Accuracy          \\ \hline
Kwapisz et al.~\cite{Kwapisz:2010} & 99.49 {[}99.28, 99.71{]} & 97.09 {[}96.72, 97.46{]} & 76.08 {[}68.27, 83.89{]} & 15.54 {[}13.11, 17.96{]} & 70.73 {[}65.32, 76.13{]}  & 52.47 {[}47.70, 57.25{]} & 84.83 {[}84.08, 85.58{]} &       70.89                \\ \hline
Catal et al.~\cite{Catal:2015}     & 99.92 {[}99.83, 1.000{]} & 97.66 {[}97.34, 97.98{]} & 91.37 {[}91.00, 91.73{]} & 48.57 {[}47.01, 50.12{]} & 82.00 {[}80.55, 83.45{]} & 67.35 {[}66.79, 67.90{]} & 90.62 {[}90.32, 90.92{]} &          82.49             \\ \hline
Kim et al.~\cite{Kim:2012}         & 99.88 {[}99.77, 99.99{]} & 94.99 {[}94.54, 95.44{]}  & 90.21 {[}89.84, 90.58{]} & 51.42 {[}50.29, 52.55{]}  & 74.97 {[}73.12, 76.82{]}  & 62.91 {[}61.41 64.41{]}  & 81.75 {[}80.71, 82.79{]} &      79.44                 \\ \hline
Chen and Xue~\cite{Chen:2015}  & 93.19 {[}91.75, 94.62{]} & 94.03 {[}93.12, 94.95 {]} & 86.78 {[}86.00, 87.55{]} & $-$                        & $-$                        & 72.92 {[}70.80, 75.03{]} & 92.65 {[}92.18, 93.12{]} &      87.91                 \\ \hline
Jiang and Yin~\cite{Jiang:2015}    & 60.69 {[}48.88, 72.50{]} &             $-$                & 82.55 {[}81.73, 83.36{]} & $-$                        & $-$                        & 72.22 {[}70.47, 73.98{]} & 91.68 {[}91.43, 91.94{]}   &        76.78               \\ \hline
Ha et al.~\cite{Ha:2015}           & 92.60 {[}91.29, 93.91{]} & 93.42 {[}92.74, 94.10{]} & $-$ & $-$                        & $-$                        & $-$                        & $-$                         &        93.00               \\ \hline
Ha and Choi~\cite{Ha:2016}         & 84.46 {[}82.42, 86.51{]} & 90.68 {[}89.93, 91.43{]} & $-$                         & $-$                        & $-$                        & $-$                        & $-$                         &      92.33                 \\ \hline
Mean Accuracy                              & 90.03                            &             94.64                &   85.39                          &                   38.51         &         75.90                   &         65.57                   &       88.30                      & \multicolumn{1}{c|}{} \\ \hline
\end{tabular}
\end{table*}
\begin{table*}[!htb]
\centering
\tiny
\caption{Mean accuracy and confidence interval using the Full-Non-Overlapping-Window and Cross Validation (FNCV) combination. The symbol '-' denotes which was not possible to execute the method on the respective dataset.}
\label{my-label2}
\begin{tabular}{|c|c|c|c|c|c|c|c|c|}
\hline
                                          & MHEALTH                     & PAMAP2                      & USCHAD                      & UTD-1                   & UTD-2                   & WHARF                       & WISDM                       & Mean Accuracy          \\ \hline
Kwapisz et al.~\cite{Kwapisz:2010} & 99.03 {[}98.41, 99.64{]} & 93.86 {[}93.03, 94.70{]} & 75.74 {[}72.06, 79.43{]} & 11.34 {[}09.42, 13.27{]} & 67.83 {[}62.21, 73.45{]} & 47.92 {[}41.52, 54.31{]} & 83.93 {[}83.49, 84.37{]} &         68.52              \\ \hline
Catal et al.~\cite{Catal:2015}     & 99.63 {[}99.32, 99.93{]} & 95.77 {[}95.23, 96.31{]} & 88.76 {[}88.12, 89.41{]} & 46.90 {[}44.56, 49.23{]} & 82.10 {[}79.28, 84.93{]} & 61.15 {[}59.35, 62.96{]} & 89.03 {[}88.48, 89.57{]} &         80.47              \\ \hline
Kim et al.~\cite{Kim:2012}         & 99.70 {[}99.40, 100.0{]} & 92.84 {[}92.31, 93.37{]}                           & 86.80 {[}85.89, 87.71{]} & 48.02 {[}46.28, 49.76{]} & 73.03 {[}69.91, 76.16{]} & 60.31 {[}58.78, 61.85{]} & 78.82 {[}77.72, 79.93{]} &         77.07              \\ \hline
Chen and Xue~\cite{Chen:2015}  & 91.61 {[}89.98, 93.25{]} & 92.36 {[}91.55, 93.18{]} & 82.63 {[}82.04, 83.23{]} & $-$                         & $-$                         & 64.77 {[}62.95, 66.59{]} & 91.47 {[}91.06, 91.88{]} &       84.56                \\ \hline
Jiang and Yin~\cite{Jiang:2015}    & 51.00 {[}29.29, 72.70{]} & $-$                         & 76.77 {[}75.78, 77.77{]} & $-$                         & $-$                         & 60.28 {[}57.37, 63.20{]} & 90.02 {[}89.47, 90.58{]} &    69.51                   \\ \hline
Ha et al.~\cite{Ha:2015}           & 90.85 {[}88.96, 92.75{]} & 90.77 {[}89.45, 92.09{]} & $-$                         & $-$                         & $-$                         & $-$                         & $-$                         &       90.81                \\ \hline
Ha and Choi~\cite{Ha:2016}         & 82.70 {[}79.76, 85.64{]} & 87.36 {[}86.20, 88.52{]} & $-$                         & $-$                         & $-$                         & $-$                         & $-$                         &        85.03               \\ \hline
Mean Accuracy                              & 87.78 & 92.16 & 82.14 & 35.41 & 74.32 & 58.85 & 86.65 & \multicolumn{1}{c|}{} \\ \hline
\end{tabular}
\end{table*}
\begin{table*}[!htb]
\centering
\tiny
\caption{Mean accuracy and confidence interval using the Leave-One-Trial-Out and Cross Validation (LTCV) combination. The symbol '-' denotes which was not possible to execute the method on the respective dataset.}
\label{tab::loto}
\begin{tabular}{|c|c|c|c|c|c|c|c|c|}
\hline
                                          & MHEALTH                  & PAMAP2                   & USCHAD                   & UTD-1                & UTD-2                & WHARF                    & WISDM                    & Mean Accuracy         \\ \hline
Kwapisz et al.~\cite{Kwapisz:2010} & 89.75 {[}85.52, 93.98{]} & 70.58 {[}64.74, 76.41{]} & 76.52 {[}73.97, 79.07{]} & 15.99 {[}13.00, 18.97{]} & 69.61 {[}63.68, 75.54{]} & 44.51 {[}36.00, 53.02{]} & 79.08 {[}76.26, 81.90{]} &          63.71             \\ \hline
Catal et al.~\cite{Catal:2015}     & 91.84 {[}87.67, 96.01{]} & 81.03 {[}75.02, 87.04{]} & 87.77 {[}86.52, 89.02{]} & 47.80 {[}45.70, 49.89{]} & 81.37 {[}78.43, 84.32{]} & 64.84 {[}63.05, 66.63{]} & 80.52 {[}76.66, 84.38{]} &         76.45              \\ \hline
Kim et al.~\cite{Kim:2012}         & 91.51 {[}87.95, 95.06{]} & 78.08 {[}71.63, 84.54{]} & 85.70 {[}84.28, 87.12{]} & 50.98 {[}50.45, 51.51{]} & 75.27 {[}72.23, 78.31{]} & 61.12 {[}58.55, 63.68{]} & 56.26 {[}52.76, 59.76{]} &              71.27         \\ \hline
Chen and Xue~\cite{Chen:2015}  & 89.95 {[}86.21, 93.70{]} & 82.32 {[}77.14, 87.50{]} & 84.66 {[}83.04, 86.29{]} & $-$                      & $-$                      & 72.55 {[}70.57, 74.54{]} & 86.55 {[}84.14, 88.96{]} &     83.20                  \\ \hline
Jiang and Yin~\cite{Jiang:2015}    & 52.78 {[}39.05, 66.52{]} & $-$                      & 80.73 {[}78.70, 82.75{]} & $-$                      & $-$                      & 70.79 {[}68.69, 72.88{]} & 83.82 {[}79.68, 87.96{]} &         72.03              \\ \hline
Ha et al.~\cite{Ha:2015}           & 85.31 {[}81.43, 89.20{]} & 80.13 {[}72.99, 87.27{]} & $-$                      & $-$                      & $-$                      & $-$                      & $-$                      &         82.71              \\ \hline
Ha and Choi~\cite{Ha:2016}         & 82.75 {[}79.23, 86.26{]} & 71.19 {[}65.70, 76.69{]} & $-$                      & $-$                      & $-$                      & $-$                      & $-$                      &        76.96               \\ \hline
Mean Accuracy                             & 83.41 & 77.22 & 83.07 & 38.25 & 75.41 & 62.76 & 77.24 & \multicolumn{1}{c|}{} \\ \hline
\end{tabular}
\end{table*}
\begin{table*}[!htb]
\centering
\tiny
\caption{Mean accuracy and confidence interval using the Semi-Non-Overlapping-Window and Leave-One-Subject-Out (SNLS) combination. The symbol '-' denotes which was not possible to execute the method on the respective dataset.}
\label{tab::loso}
\begin{tabular}{|c|c|c|c|c|c|c|c|c|}
\hline
                                          & MHEALTH                     & PAMAP2                      & USCHAD                      & UTD-1                   & UTD-2                   & WHARF                       & WISDM                       & Mean Accuracy         \\ \hline
Kwapisz et al.~\cite{Kwapisz:2010} & 90.41 {[}86.54, 94.28{]} & 71.27 {[}52.07, 90.47{]} & 70.15 {[}65.06, 75.24{]} & 13.04 {[}10.19, 15.90{]} & 66.67 {[}59.21, 74.14{]} & 42.19 {[}33.40, 50.98{]} & 75.31 {[}70.07, 80.55{]} &        61.29               \\ \hline
Catal et al.~\cite{Catal:2015}     & 94.66 {[}91.17, 98.15{]} & 85.25 {[}76.27, 94.22{]} & 75.89 {[}71.62, 80.16{]} & 32.45 {[}30.18, 34.71{]} & 74.67 {[}65.75, 83.58{]} & 46.84 {[}41.02, 52.67{]} & 74.96 {[}69.66, 80.27{]} &         69.29              \\ \hline
Kim et al.~\cite{Kim:2012}         & 93.90 {[}90.05, 97.74{]} & 81.57 {[}73.50, 89.64{]} & 64.20 {[}58.88, 69.53{]} & 38.05 {[}37.01, 39.09{]} & 64.60 {[}59.73, 69.47{]} & 51.48 {[}46.11, 56.84{]}                           & 50.22 {[}45.85, 54.59{]} &          63.43             \\ \hline
Chen and Xue~\cite{Chen:2015}  & 88.67 {[}85.38, 91.96{]} & 83.06 {[}75.40, 90.71{]} & 75.58 {[}70.05, 81.11{]} & $-$                         & $-$                         & 61.94 {[}55.02, 68.86{]} & 83.89 {[}79.72, 88.06{]} &           78.62            \\ \hline
Jiang and Yin~\cite{Jiang:2015}    & 51.46 {[}35.35, 67.57{]} & $-$                         & 74.88 {[}71.28, 78.48{]} & $-$                         & $-$                         & 65.35 {[}58.81, 71.88{]} & 79.97 {[}74.21, 85.73{]} &        67.91               \\ \hline
Ha et al.~\cite{Ha:2015}           & 88.34 {[}84.91, 91.78{]} & 73.79 {[}59.56, 88.02{]} & $-$                         & $-$                         & $-$                         & $-$                         & $-$                         &       81.06                \\ \hline
Ha and Choi~\cite{Ha:2016}         & 84.23 {[}80.01, 88.44{]} & 74.21 {[}60.91, 87.50{]} & $-$                         & $-$                         & $-$                         & $-$                         & $-$                         &         79.21              \\ \hline
Mean Accuracy                             &  84.52 &  78.19   & 72.14  & 27.84  & 68.64  & 53.55 & 72.87  & \multicolumn{1}{c|}{} \\ \hline
\end{tabular}
\end{table*}

Regarding the methods based on ConvNets, the approach with the best performance is the method of Chen and Xue~\cite{Chen:2015}, achieving the Top1 one and three times when evaluated on LTCV and SNLS, respectively. This result is a consequence of the filter shapes, which are adequate to capture the temporal and spatial pattern of the signal. On the other hand, the methods of Jiang and Yin~\cite{Jiang:2015}, Ha et al.~\cite{Ha:2015} and, Ha and Choi~\cite{Ha:2016} were not able to achieve good results. We believe that these inaccurate results are an effect of their convolutional filters ($3\times3 $ and $5\times5$), which capture a small temporal pattern besides being sensitive to noise. 
In particular, the methods~\cite{Ha:2015} and \cite{Ha:2016} were evaluated only on two datasets (due to issues of the network architecture). However, by normalizing their results by the number of datasets evaluated, they still do not present good performance. Finally, by considering handcrafted and ConvNets approaches, the more accurate methods are the approaches of Catal et al.~\cite{Catal:2015} and Chen and Xue~\cite{Chen:2015}. This result indicates that, though ConvNets-based approaches have presented remarkable results in human activity recognition based on wearable data, handcrafted approaches are able to achieve comparable results. 
\section{Conclusions}\label{sec::conclusion}
This work conducted an extensive set of experiments to demonstrate essential issues which currently are not considered during the evaluation of the human activity recognition based on wearable sensor data. The main issue is regarding the process employed to generate the data samples, where the traditional process is susceptible to bias leading to skewed results.  To demonstrate this, we investigate novel techniques to generate the data samples, which focus on reducing and removing this bias. According to our experiments, the accuracy drops considerably when appropriated data generation processes (bias-invariant) are used. Hence, the results reported by previous works can be skewed and do not reflect their real performance. In addition, throughout the experiments, we implement several top-performance methods and evaluated them on many popular and publicity available datasets. Thereby, we define the state-of-the-art methods in human activity recognition based on wearable sensor data.

We highlight that, different from previous studies and surveys, our work does not summarize or discuss existing methods based on their reported results, which makes this work, to the best of our knowledge, the first that implements, groups and handles important issues regarding the activity recognition associated with wearable sensor data.
\section*{Acknowledgments}
The authors would like to thank the Brazilian National Research Council - CNPq (Grant \#311053/2016-5), the {Minas} Gerais Research Foundation - FAPEMIG (Grants APQ-00567-14 and PPM-00540-17) and the Coordination for the Improvement of Higher Education Personnel -- CAPES ({DeepEyes} Project).
Part of the results presented in this paper were obtained through research on a project titled " HAR-HEALTH: Reconhecimento de Atividades Humanas associadas a Doenças Cr\^{o}nicas ", sponsored by Samsung Eletr\^{o}nica da Amaz\^{o}nia Ltda. under the terms of Brazilian federal law No. 8.248/91. 

\balance
\bibliographystyle{IEEEbib}
\bibliography{References}

%

%
\end{document}